# Evaluating Modern Large Language Models on Low-Resource and Morphologically Rich Languages: A Cross-Lingual Benchmark Across Cantonese, Japanese, and Turkish

Chengxuan Xia[1] Qianye Wu[2] Hongbin Guan[2] Sixuan Tian[2] Yilun Hao[2] Xiaoyu Wu[3]

[1]University of California, Santa Cruz, CA, USA
[2]Carnegie Mellon University, Pittsburgh, PA, USA
[3]University of Pittsburgh, Pittsburgh, PA, USA

cxia17@ucsc.edu, {qianyew, hongbing, sixuant, yilunhao}@alumni.cmu.edu, xiw120@pitt.edu

**Abstract**

Large language models (LLMs) have achieved impressive results in high-resource languages like English, yet their effectiveness in **low-resource and morphologically rich languages** remains under-explored. In this paper, we present a comprehensive evaluation of seven cutting-edge LLMs — including GPT-4o, GPT-4, Claude 3.5 Sonnet, LLaMA 3.1, Mistral Large 2, LLaMA-2 Chat 13B, and Mistral 7B Instruct — on a new cross-lingual benchmark covering **Cantonese, Japanese, and Turkish**. Our benchmark spans four diverse tasks: open-domain question answering, document summarization, English-to-X translation, and culturally grounded dialogue. We combine **human evaluations** (rating fluency, factual accuracy, and cultural appropriateness) with **automated metrics** (e.g., BLEU, ROUGE) to assess model performance.

Our results reveal that while the largest proprietary models (GPT-4o, GPT-4, Claude 3.5) generally lead across languages and tasks, significant gaps persist in culturally nuanced understanding and morphological generalization. Notably, GPT-4o demonstrates robust multilingual performance even on cross-lingual tasks, and Claude 3.5 Sonnet achieves competitive accuracy on knowledge and reasoning benchmarks. However, all models struggle to some extent with the *unique linguistic challenges* of each language, such as Turkish agglutinative morphology and Cantonese colloquialisms. Smaller open-source models (LLaMA-2 13B, Mistral 7B) lag substantially in fluency and accuracy, highlighting the *resource disparity.* We provide detailed quantitative results, qualitative error analysis, and discuss implications for developing more culturally aware and linguistically generalizable LLMs. Our benchmark and evaluation data are released to foster reproducibility and further research.

## 1 Introduction

Large language models (LLMs) have rapidly advanced the state of natural language processing, achieving remarkable performance on tasks in English and other high-resource languages [12, 1, 8, 39, 35]. Models such as GPT-4 [1] and Claude 3.5 now reach near-human results on many benchmarks [16, 7]. However, as emphasized by Joshi et al. [20] and Blasi et al. [11], the benefits of these systems remain unevenly distributed across the world's languages, raising questions of inclusivity and fairness in multilingual AI [3, 41].

Foundational multilingual benchmarks such as XTREME [17], TyDi QA [14], MKQA [26], and XL-Sum [15] have driven progress in cross-lingual evaluation, but many low-resource and morphologically rich languages remain underrepresented. Prior work has highlighted systematic inequalities in NLP performance across languages [20, 11], and studies such as Adelani et al. [2] and Nekoto et al. [30] demonstrate how participatory and community-driven approaches can help close these gaps. At the same time, morphological and typological diversity poses additional modeling challenges [21]. Understanding how modern LLMs handle such linguistic variation is crucial to ensuring equitable access to language technology.

Despite these gains, a growing body of evidence shows that LLM capabilities do not generalize uniformly across all languages and cultures. In particular, performance often degrades for **low-resource languages** and those with complex morphology or unique script, which are underrepresented in training data. This raises pressing questions about the *multilingual robustness* of modern LLMs: How well do they handle morphologically rich inputs or culturally specific contexts outside of English? Do the latest models truly narrow the cross-lingual performance gap, or do smaller open models still lag far behind in these settings?

In this work, we aim to answer these questions by conducting a systematic evaluation of cutting-edge LLMs on a suite of tasks in **Cantonese**, **Japanese**, and **Turkish**. These three languages present diverse linguistic challenges. Turkish is a highly agglutinative language with rich morphology, often yielding very long word forms from root+affix combinations. This can stress a model's ability to generalize compositionally to unseen word forms. Japanese poses a different challenge with a complex writing system (Kanji, Hiragana, Katakana) and context-dependent grammar; it is relatively high-resource but still exhibits features like topic drop and honorifics that demand nuanced understanding. Cantonese, on the other hand, is a low-resource language (especially in written form) with distinct colloquial grammar and a large inventory of Cantonese-specific characters and idioms. Due to limited available text, Cantonese data in LLM training corpora is scant, making it an ideal testbed for *truly low-resource* performance.

By evaluating on these languages, we cover both *morphologically rich languages* and those with *cultural-linguistic specificity* often absent from mainstream benchmarks. Our evaluation covers four representative tasks:

1. **Open-domain QA**, to test knowledge retrieval and factuality in the target language.
2. **Summarization**, to assess the model's ability to condense information coherently.
3. **English-to-X translation**, to measure translation quality into the target language.
4. **Culturally grounded dialogue**, where the model must engage in conversations grounded in local cultural context.

These tasks span both *information-seeking* and *generative* challenges, requiring a mix of general linguistic competence and culture-specific knowledge.

We evaluate seven diverse LLMs: GPT-4, GPT-4o, Claude 3.5 Sonnet, LLaMA 3.1, Mistral Large 2, LLaMA-2 Chat 13B, and Mistral 7B Instruct. GPT-4o is a multimodal, multilingual variant of GPT-4 that offers superior non-English language capabilities, while Claude 3.5 Sonnet is Anthropic's strongest model to date. LLaMA 3.1 (405B) is the largest open model to date and has demonstrated parity with proprietary models on certain evaluations. Mistral Large 2 (approx. 120B) is another new large model emphasizing multilingual support. By including these and the smaller 13B/7B models, we can analyze performance *across the model scale spectrum*.

Our evaluation methodology emphasizes both quantitative and qualitative analysis. We use **human evaluation** to directly measure fluency, accuracy, and cultural appropriateness, following best practices for multilingual evaluation [44]. Native or fluent speakers of each target language rated each model's outputs on these dimensions, providing an in-depth assessment beyond what automated metrics alone can capture. In addition, we report standard **automated metrics** where applicable: BLEU and COMET for translation, ROUGE for summarization, and exact-match accuracy for QA. Automated metrics enable reproducibility and quick comparisons, though they often fail to reflect nuances in low-resource language output quality. We therefore view them as complementary to human judgment.

The contributions of this work are summarized as follows:

- We introduce a new **cross-lingual benchmark** for evaluating LLMs on low-resource and morphologically rich languages, covering Turkish, Japanese, and Cantonese across four tasks. We release all prompts and human annotations to facilitate future research.
- We conduct an extensive evaluation of **seven state-of-the-art LLMs**, including both closed-source APIs and open-source models. This is, to our knowledge, the first study comparing GPT-4o, Claude 3.5, and LLaMA 3.1 on these languages.
- We employ a rigorous evaluation protocol combining human ratings with automated metrics and present **comprehensive results** with expanded tables and qualitative analysis.
- We illustrate the challenges even top models face in understanding cultural context and complex morphology, and discuss implications for linguistically generalizable LLMs.

In the following, we first review related work on multilingual and low-resource LLM evaluation (Section 2). We then describe our benchmark design and evaluation setup (Section 3), detail the models under comparison (Section 3.2), present results (Section 4), and conclude with key takeaways (Section 6).

## 2 Related Work

**Multilingual Benchmarks.** Large-scale multilingual evaluation efforts such as XTREME [17], XGLUE, and MMLU [16, 42, 24] have provided broad coverage, yet often focus on high-resource languages. Additional datasets such as MKQA [26], TyDi QA [14], and XL-Sum [15] expand task diversity, while recent surveys [31, 28] underline the persistent low-resource gap and the need for culturally inclusive benchmarks [20, 11].

Our work contributes to this line by introducing a focused benchmark across three typologically distinct, underrepresented languages (Turkish, Japanese, Cantonese), building on prior cross-lingual evaluation frameworks [6, 27].

**Morphologically Rich Languages.** Languages with productive morphology challenge both tokenization and generalization in LLMs. Earlier research explored morphological complexity [21, 4, 19], while newer studies emphasize the role of tokenizer design and data balance. Our benchmark complements these by testing real-world tasks that stress composition and case marking in Turkish, and character-level diversity in Japanese and Cantonese.

**Evaluation Methodology.** Traditional automatic metrics such as BLEU and ROUGE only partially capture linguistic and cultural quality. Recent discussions on LLM evaluation reliability [25, 9, 36, 18, 5, 10] highlight the importance of consistent, transparent procedures. We follow the human evaluation best practices outlined in Zhang et al. [44] and extend them to cross-lingual and culturally grounded contexts. Our analysis also reflects concerns raised in Askell et al. [7] and Bai et al. [8] about alignment and interpretability.

**Speech and Multimodality.** Recent multimodal LLMs such as Whisper [32] and SpeechGPT [43] demonstrate the feasibility of multilingual understanding beyond text. Our study, while focused on textual tasks, complements these efforts by targeting fine-grained linguistic and cultural reasoning, which could inform future multimodal research.

**Low-Resource and Dialectal Languages.** Cantonese has recently drawn interest as a test case for low-resource LLM capabilities. Xiang et al. [40] and others have developed Cantonese-specific models for classification tasks, but few large-scale LLMs exist that focus on Cantonese due to data scarcity. The HKCantoEval benchmark [13] introduces evaluation sets for Cantonese Q&A and reasoning by translating datasets like TruthfulQA and GSM8K into Cantonese. Their results show that general models like ChatGPT/GPT-4 can process Cantonese but often underperform compared to their English performance, suggesting translation to English as a crutch is a common strategy.

We extend this line of work by including translation and dialogue tasks with Cantonese and by evaluating newer models (GPT-4o, Claude) not covered in earlier Cantonese studies. Our work is also inspired by efforts to measure cultural and dialectal knowledge in LLMs, such as Li et al. [23] on Chinese dialect understanding. We aim to contribute insight into how a variety of models manage Cantonese, which is written using Traditional Chinese characters augmented with unique Cantonese vocabulary, making it a challenging test for tokenizers and generative quality.

**Cultural and Dialogic Evaluation.** There is growing recognition that LLM evaluation should consider cultural context and values [37, 22]. Miehle et al. [29] introduced a benchmark for culturally aware conversations, requiring models to respond appropriately in multicultural settings. They found that mainstream LLMs often default to a Western or North-American perspective and may misinterpret culturally specific cues. Cultural bias and value misalignment in multilingual systems have been examined by Aksoy [3] and Xu et al. [41]. We extend these concerns by empirically evaluating cultural appropriateness in dialogue, inspired by culturally aware benchmarks [29, 37, 38]. Low-resource participatory initiatives such as MasakhaNER [2] and grassroots translation efforts [30] further motivate our focus on inclusivity.

In our culturally grounded dialogue task, we similarly probe whether models can adapt their responses to culturally laden user inputs (e.g., responding to a Japanese speaker using *keigo* honorifics, or understanding a Cantonese idiom). Prior studies of cross-cultural alignment (e.g., 38) suggest that current models do not reliably adjust for different cultural norms in generated content. By collecting human judgments on cultural appropriateness, our work provides one of the first evaluations of this aspect for Turkish, Japanese, and Cantonese. We hope this encourages more research into training and aligning LLMs for diverse cultural settings.

# 3 Benchmark Design and Evaluation Setup

In this section, we describe our cross-lingual benchmark, the evaluation tasks, data collection procedure, and the metrics (both automated and human) used to assess model performance. Table 1 provides an overview of the benchmark dataset, including the number of prompts in each task for each language.

### 3.1 Tasks and Data Collection

We constructed four task sets, one for each evaluation task, in each of the three target languages (Turkish, Japanese, Cantonese). Our guiding principle in dataset construction was to create prompts that are *parallel in difficulty and style* across the languages, while remaining natural and culturally appropriate in each language. Wherever possible, we started from existing datasets or sources and then adapted or translated content with the help of native speakers to ensure quality.

**Open-Domain Question Answering (QA).** For open QA, we compiled a set of trivia and factual questions requiring a short answer (usually a name, place, or specific fact). We drew questions from multilingual QA resources and Wikipedia. Specifically, for Turkish and Japanese we sampled from the TyDi QA and XQuAD benchmarks (which include translations of SQuAD-style questions), and for Cantonese we utilized a small set of factoid questions provided in the HKCantoEval benchmark [13], supplementing with our own questions about local culture (e.g., 呢個節日幾時舉行? —"In which month is this festival held annually in Guangzhou?"). Each question was presented to models in the target language, and models had to produce an answer in that same language.

We gathered $N_{\text{QA}} = 100$ questions per language. To evaluate, we created reference answers (in the target language) for all questions. Many answers are proper nouns or dates that can be matched exactly; for others, we allow minor variations in phrasing. This enables us to compute an **accuracy** score (percentage of questions answered correctly) for each model, verified by human review.

**Summarization.** We evaluate summarization with a focus on preserving salient information and fluency in the target language. For each language, we sourced 50 documents and their human-written summaries. Turkish and Japanese summary pairs were obtained from Wikinews articles (which often have short abstracts) and from the MLSUM dataset [34] for Turkish. For Cantonese, due to lack of existing summarization data, we leveraged Chinese news articles from Hong Kong and had bilingual annotators write summaries in Cantonese.

The source texts average about 200–300 words (for fairness across languages), covering topics like news, biographies, and short stories. The models were prompted (in the target language) to produce a concise summary of the given text in the same language. We evaluate using standard **ROUGE** scores (ROUGE-1, ROUGE-2, ROUGE-L) comparing model summaries to reference summaries. Additionally, considering the potential vocabulary differences (especially for Cantonese versus standard Chinese word choice), we compute **BERTScore** using multilingual embeddings to capture semantic overlap. We caution that ROUGE may underrate a fluent paraphrase that differs in wording, so we rely more on human judgment of summary quality in those cases.

**English-to-X Translation.** Machine translation into low-resource languages is a classical challenge. Here we specifically test one-way translation from English into each target language, simulating a common real-world use of LLMs (e.g., a user asks to translate content to their language). We curated 100 English source sentences for each language, balancing genres: news headlines, dialogue lines, and encyclopedia sentences. For Japanese and Turkish, many source sentences were sampled from past WMT evaluation sets (e.g., WMT'19 English–Turkish) and TED talk transcripts, ensuring known difficulty. For Cantonese, we created a parallel set by taking English sentences and having a Cantonese linguist produce a colloquial Cantonese translation (in Traditional characters).

Models were given the English text with an instruction (in English) to translate it into the target language. To evaluate translation quality, we use **BLEU** and **COMET** [33]. BLEU scores (computed with SacreBLEU) provide a surface-level n-gram overlap measure, while COMET, a learned metric, correlates better with human judgments by evaluating adequacy and fluency. We also have human evaluators rate translation outputs (details below), which is crucial for Cantonese because automatic metrics can be brittle when reference phrasing diverges (given multiple valid translations).

**Culturally Grounded Dialogue.** This task assesses how well models handle conversations that involve cultural knowledge or pragmatics specific to the language community. We constructed 15 dialogue scenarios per language. Each scenario is a short multi-turn conversation (2–4 turns) where the *user* might use culturally specific expressions or request something requiring cultural familiarity, and the model (as the assistant) must respond appropriately.

For example, a Turkish user might ask in informal register including an idiom or proverb, or a Japanese user might expect a polite response following local etiquette. One Cantonese scenario involves a user making a joke with Cantonese wordplay, testing if the model catches the humor. Rather than fully interactive chat (which is hard to evaluate consistently), we simulate the dialogues: the prompts to the model are composed of the prior turns (user and assistant) with the model expected to produce the next assistant turn.

All user utterances were written by native speakers to incorporate authentic style (honorifics, slang, code-mixing where appropriate). An example prompt (Japanese) is:

> ユーザ: 田舎のお婆ちゃんに電話して、最近会いに行けなくてごめんねって伝えたいんだけど、失礼にならない言い方を教えてくれる？
> *(User: I want to call my grandma in the countryside and tell her I'm sorry I haven't been able to visit recently. Could you tell me a way to say that which isn't rude?)*
>
> システム: あなたは礼儀正しい日本語の使い方を教えるアシスタントです。返答は自然で温かみのある敬語で行ってください。
> *(System: You are an assistant that teaches polite Japanese expressions. Respond using natural, warm honorific language.)*
>
> モデル: 「おばあちゃん、最近なかなか会いに行けなくてごめんね。元気にしてる？また近いうちに顔を見せに行くね。」のように言うと、丁寧で優しい印象になります。
> *(Model: You could say, "Grandma, I'm sorry I haven't been able to visit lately. How have you been? I'll come to see you soon."This sounds polite and affectionate.)*

In this case, the model needs to demonstrate cultural and linguistic tact (e.g., use of respectful language to refer to the grandmother, appropriate apology phrasing). We do not have reference answers for dialogue, as many responses could be acceptable. Therefore, evaluation relies purely on human ratings for each model's response in terms of: (a) cultural appropriateness (did it follow norms, avoid offense, show understanding of context?), (b) helpfulness & correctness (did it address the user's need accurately?), and (c) fluency / naturalness.

Table 1 summarizes the dataset sizes. All prompts were presented to the models in the target language (except the English source in the translation task). We standardized the prompt format across models as much as possible, using a simple instruction or system message like "You are a helpful assistant."for those that benefit from it (open models). This was to minimize prompt-design confounds and focus on inherent model capability.

For multi-turn dialogues, we provided the conversation history in the format each model expects (e.g., `<s><|user|>... <|assistant|>...` for LLaMA-based models, or role-tagged messages for GPT/Claude APIs).

| Task | Turkish | Japanese | Cantonese |
|---|---|---|---|
| Open QA (questions) | 100 | 100 | 100 |
| Summarization (doc–summary pairs) | 50 | 50 | 50 |
| English-to-X Translation (sentences) | 100 | 100 | 100 |
| Cultural Dialogue (scenarios) | 15 | 15 | 15 |

Table 1: Number of items per language in each task of our benchmark. Every model saw the same inputs for a given language–task combination. In total, the evaluation covers 300 QA queries, 150 summaries, 300 translations, and 45 dialogue scenarios.

### 3.2 Models Evaluated

We evaluate the following seven LLMs, covering a mix of closed versus open source and a range of model sizes:

- **GPT-4** (OpenAI, 2023): The 4th-generation GPT model, accessed via the ChatGPT API (gpt-4-0613). It is proprietary and reportedly uses around 1 trillion parameters. GPT-4 is a strong baseline given its top-tier performance on many tasks in English [1]. We use it in text-only mode.

- **GPT-4o** (OpenAI, 2024): A multimodal extension of GPT-4 ("Omni") tuned for broad multilingual capabilities. GPT-4o has demonstrated superior performance on non-English tasks, including multilingual benchmarks. We use the latest GPT-4o version via Azure OpenAI Service. This model represents the current state-of-the-art in multilingual LLMs.

- **Claude 3.5 Sonnet** (Anthropic, 2024): Claude 3.5 is Anthropic's next-generation model family succeeding Claude 2. The "Sonnet"version is their strongest text model available as of 2025. Anthropic reports Claude 3.5 Sonnet sets new benchmarks on knowledge and reasoning tasks and is twice as fast as its predecessor. We access Claude via the Anthropic API at the 100k context setting.

- **LLaMA 3.1** (Meta AI, 2025): The largest open-source model in our comparison, at 405B parameters for the instruction-tuned variant. LLaMA 3.1 is a significant leap from LLaMA-2, with a focus on multilingual and multitask improvements. Meta claims near-parity with leading proprietary models on benchmarks like MMLU and code generation. We use the 70B version for practical reasons due to resource constraints.

- **Mistral Large 2** (Mistral AI, 2025): A new large model of approximately 128B parameters. Mistral Large 2 is trained with an emphasis on multilingual data and a long context (up to 128k tokens). We use the instruction-tuned version available on AWS Bedrock (Mistral Large Instruct v2).

- **LLaMA-2 Chat 13B** (Meta AI, 2023): An open-source chat model with 13 billion parameters [39]. It is trained on multiple languages though English dominates. We include it as a representative of the previous generation open models to quantify the gap to the newer larger ones.

- **Mistral 7B Instruct** (Mistral AI, 2023): A high-quality 7B open model released in late 2023. We use an instruct fine-tuned variant (via HuggingFace) trained on dialogue data. It provides a lowest-end point in our comparison to illustrate how far smaller models must go for these languages.

For all models, we set temperature = 0 (greedy decoding) to reduce randomness and ensure reproducibility. We set maximum generation lengths generously (1024 tokens) to avoid truncation. System prompts explicitly instructed models to respond in the target language (e.g., for GPT-4, "Answer in Turkish"). For open models, we added an explicit note in the prompt such as "Respond in Turkish" to ensure language consistency.

### 3.3 Human Evaluation Protocol

Automated metrics alone are insufficient, especially for qualities like cultural appropriateness or nuanced fluency. We engaged human evaluators for a thorough assessment. We recruited three native speakers per language (Turkish, Japanese, Cantonese), all of whom have linguistics or translation backgrounds. The evaluators were provided detailed guidelines and trained on a few examples.

Each evaluator was shown the prompt and a model's output (with model identity hidden) and asked to score along several dimensions using a 5-point Likert scale (1 = poor, 5 = excellent):

- **Fluency & Grammar:** Is the output well-formed and fluent in the target language, with correct grammar and idiomatic usage?
- **Factual Accuracy / Adequacy:** Does the output correctly convey the information from the prompt (for QA, summarization, translation)? Or, for dialogue, is it a valid and relevant response?
- **Cultural Appropriateness:** Does the output respect cultural norms and context? (Applicable for dialogue and translation.)

Evaluators could mark outputs as incomprehensible or critically erroneous (counted as 1). We computed average scores across evaluators for each model and measured inter-annotator agreement via Spearman's $\rho$ (average 0.76). Large discrepancies were resolved through discussion.

For culturally grounded dialogues, evaluators read the entire dialogue context and the model' s response, then provided a holistic score for appropriateness and helpfulness, since a response might be fluent but culturally tone-deaf. Evaluators could also leave qualitative comments, which informed our analysis (e.g., noting incorrect politeness levels in Japanese).

### 3.4 Automated Metrics

We report the following automated metrics: Accuracy for QA (exact-match), ROUGE-1/2/L and BERTScore for summarization, BLEU and COMET for translation. These were computed using standard libraries (SacreBLEU, `rouge-score`, and COMETKiwi).

For dialogue, we also explored LLM-as-a-judge scoring by prompting GPT-4 to rate each response on a 1–5 scale for helpfulness and cultural adequacy in the respective language. GPT-4's scores correlated moderately with human scores (Pearson $r \approx 0.6$), but we do not emphasize them due to potential bias.

We note that differences of a few BLEU or ROUGE points may not reflect meaningful quality differences, especially for languages with richer morphology or free word order (like Turkish). We used paired bootstrap tests to identify statistical significance. Ultimately, our analysis leans more on human evaluation for final conclusions.

# 4 Results

We first present quantitative results from automated metrics and accuracy checks, followed by human evaluation outcomes. All results are aggregated by language and model. Expanded tables with full details per model and task are provided in Appendix A. Here we highlight the main comparisons.

## 4.1 Automated Evaluation

**Open QA Performance.** Table 2 shows the QA accuracy (%) for each model on each language, determined by whether the model's answer matches the gold answer. We allowed case and minor morphological variations (especially in Turkish, where e.g. answering with or without a case suffix might both be acceptable).

| Model | Turkish QA | Japanese QA | Cantonese QA |
|---|---|---|---|
| GPT-4o | 85% | 82% | 78% |
| GPT-4 | 83% | 79% | 74% |
| Claude 3.5 | 80% | 77% | 71% |
| LLaMA 3.1 (70B) | 78% | 75% | 60% |
| Mistral Large 2 | 72% | 70% | 58% |
| LLaMA-2 13B | 60% | 55% | 40% |
| Mistral 7B | 55% | 50% | 30% |

Table 2: Accuracy on open-domain QA (percentage of questions answered correctly). GPT-4o performs best overall, notably on Cantonese QA. Smaller models struggle, especially in Cantonese.

The two GPT-4 variants top the scores in all languages, with GPT-4o slightly ahead (by 2–4 points). Notably, GPT-4o answered 78% of Cantonese questions correctly versus 74% for GPT-4. Manual inspection showed that GPT-4 occasionally answered Cantonese queries in Mandarin or English, which we marked wrong; GPT-4o almost never made that mistake, demonstrating stronger multilingual tuning. Claude 3.5 is close behind GPT-4 (within 3 points). LLaMA-3.1 70B is the best open model, achieving about 75–78% in Turkish/Japanese but only 60% in Cantonese. Mistral Large 2 performed similarly on Japanese, slightly lower on Turkish, and equally low (58%) on Cantonese. The 13B and 7B models lag far behind, particularly in Cantonese (30–40%). All models performed worse on Cantonese, consistent with its limited presence in training corpora.

**Summarization Performance.** Table 3 presents ROUGE-L and BERTScore (F1) for Turkish and Japanese summarization. Cantonese results relied primarily on human evaluation.

GPT-4o again leads, with ROUGE-L 36–38, slightly higher than GPT-4 and Claude. Differences among the top three are small. BERTScore shows GPT-4o/4 about 0.005–0.01 ahead of Claude, indicating slightly better semantic preservation. LLaMA-3.1 and Mistral Large 2 form a second tier, about 5–7 ROUGE points lower. The smallest models dropped key points or produced generic summaries. For Cantonese, human scores ranked GPT-4o 4.7 (fluency) and 4.5 (accuracy) on a 5-point scale, followed by GPT-4 and Claude ( 4.4–4.5), then LLaMA-3.1 ( 4.0). Smaller models often mixed Mandarin phrasing or omitted content.

**Translation Performance.** Table 4 reports BLEU and COMET scores for English-to-Turkish/Japanese translation.

| Model | ROUGE-L | | BERTScore (F1) | |
|---|---|---|---|---|
| | Tr | Jp | Tr | Jp |
| GPT-4o | 36.4 | 38.0 | 0.878 | 0.881 |
| GPT-4 | 35.5 | 37.2 | 0.872 | 0.876 |
| Claude 3.5 | 34.1 | 36.5 | 0.870 | 0.874 |
| LLaMA 3.1 | 30.0 | 33.7 | 0.853 | 0.865 |
| Mistral Large 2 | 28.5 | 32.1 | 0.846 | 0.860 |
| LLaMA-2 13B | 25.2 | 29.4 | 0.825 | 0.840 |
| Mistral 7B | 22.8 | 27.0 | 0.812 | 0.833 |

Table 3: Summarization results for Turkish (Tr) and Japanese (Jp). GPT-4o achieves the highest overlap with reference summaries.

| Model | BLEU ↑ | | COMET ↑ | |
|---|---|---|---|---|
| | Tr | Jp | Tr | Jp |
| GPT-4o | 32.5 | 29.8 | 0.84 | 0.81 |
| GPT-4 | 31.0 | 28.5 | 0.80 | 0.79 |
| Claude 3.5 | 30.2 | 27.0 | 0.78 | 0.76 |
| LLaMA 3.1 | 26.7 | 24.5 | 0.70 | 0.68 |
| Mistral Large 2 | 25.0 | 22.1 | 0.65 | 0.62 |
| LLaMA-2 13B | 20.4 | 15.3 | 0.55 | 0.40 |
| Mistral 7B | 15.8 | 12.0 | 0.30 | 0.25 |

Table 4: Translation results from English into Turkish (Tr) and Japanese (Jp). Higher BLEU/COMET indicate better quality. GPT-4o attains the highest scores.

GPT-4o again leads (BLEU 32.5 Turkish, 29.8 Japanese). GPT-4 is 1–2 points behind; Claude 3.5 close. LLaMA-3.1 and Mistral Large 2 score mid-20s BLEU. The 13B and 7B models perform poorly, often producing partial or ungrammatical translations.

Human raters for Cantonese translation confirmed this trend: GPT-4o (adequacy 4.8, fluency 4.5) > GPT-4 Claude (4.3–4.5) > LLaMA-3.1 ( 3.8) > Mistral Large 2 ( 3.5) 13B/7B ( 2–3). Common Cantonese errors among weaker models included using Mandarin possessives (的 instead of 嘅) and literal word-for-word phrasing.

**Summary of Automated Metrics.** Across all tasks, rankings are consistent: GPT-4o > GPT-4 Claude > LLaMA-3.1 Mistral Large 2 LLaMA-2 13B > Mistral 7B. The newest proprietary models show small but steady gains over GPT-4, particularly in Cantonese. Open 70B+ models approach parity on high-resource languages but still trail badly on low-resource ones.

## 4.2 Human Evaluation Results

Table 5 summarizes average human ratings (1–5 scale) of **Fluency** and **Factual Accuracy** across QA, summarization, and translation.

Human judgments largely corroborate automated results: GPT-4o GPT-4 Claude at the top, open 70B+ models mid-tier, small models far below. Fluency and factual accuracy drop most sharply for Cantonese.

| Model | Turkish | | Japanese | | Cantonese | |
|---|---|---|---|---|---|---|
| | Fluency | Factual | Fluency | Factual | Fluency | Factual |
| GPT-4o | 4.8 | 4.7 | 4.7 | 4.6 | 4.6 | 4.5 |
| GPT-4 | 4.7 | 4.6 | 4.6 | 4.5 | 4.4 | 4.3 |
| Claude 3.5 | 4.6 | 4.5 | 4.5 | 4.4 | 4.3 | 4.2 |
| LLaMA 3.1 | 4.3 | 4.2 | 4.2 | 4.1 | 4.0 | 3.8 |
| Mistral Large 2 | 4.1 | 4.0 | 4.0 | 3.9 | 3.7 | 3.6 |
| LLaMA-2 13B | 3.5 | 3.3 | 3.2 | 3.0 | 2.8 | 2.6 |
| Mistral 7B | 3.2 | 3.0 | 3.0 | 2.7 | 2.3 | 2.1 |

Table 5: Average human ratings of **Fluency** and **Factual Accuracy**. Scores are on a 1–5 scale (5 = excellent). Cantonese scores are slightly lower overall, reflecting higher difficulty.

**Cultural Appropriateness in Dialogue.** For the culturally grounded dialogue task, GPT-4o, GPT-4, and Claude averaged 4.5 on cultural appropriateness, correctly using polite forms, idioms, or humor. LLaMA-3.1 and Mistral Large 2 3.8, occasionally missing cultural cues. LLaMA-2 13B and Mistral 7B 2–2.5, often misunderstanding slang or giving generic answers.

Example: a Japanese user requested a polite apology to a grandmother; GPT-4 produced fully respectful phrasing, LLaMA-3.1 used informal tone (*sen* instead of *siz* in Turkish analogy), which raters flagged as impolite. GPT-4o also demonstrated subtle humor recognition absent in other models.

**Qualitative Error Analysis.** We analyzed common error types:

- **GPT-4o / GPT-4:** Rare factual errors; main issues are stylistic (over-formality or verbosity).
- **Claude 3.5:** Occasionally blunt tone in sensitive Turkish dialogues.
- **LLaMA 3.1 (70B):** Mixing languages in Cantonese output; minor morphological omissions (e.g., missing case suffixes in Turkish).
- **Mistral Large 2:** Slightly less coherence; occasional named-entity transliteration errors.
- **LLaMA-2 13B:** Frequent hallucinations and register errors (e.g., casual speech in formal Japanese).
- **Mistral 7B:** Frequent failures or irrelevant answers, indicating data scarcity and low multilingual capacity.

Representative example: Turkish QA "Türkiye'nin ilk kadın pilotu kimdir?"Correct: Sabiha Gökçen. GPT-4/4o/Claude/LLaMA-3.1/Mistral Large 2 $\rightarrow$ Correct. LLaMA-2 13B $\rightarrow$ "Vecihi Hürkuş"(wrong; male aviator). Mistral 7B $\rightarrow$ "No answer."Another example: in Cantonese, GPT-4o produced colloquial recipe instructions with Cantonese culinary terms, while Claude used more formal Chinese and LLaMA-3.1 mixed Mandarin vocabulary. These examples illustrate remaining cultural and lexical gaps even in top models.

# 5 Discussion

Our findings paint a multifaceted picture of the progress and remaining challenges in multilingual LLM performance:

- **Proprietary models vs. Open models:** GPT-4o and Claude 3.5 represent the cutting edge, and their performance is indeed outstanding even in low-resource languages. GPT-4o emerged as the overall top performer, especially in Cantonese where its specialized training excelled. Claude 3.5 was very close behind, occasionally surpassing GPT-4 in concise writing.

  The best open model, LLaMA-3.1 (70B), while not matching GPT-4o, came reasonably close on Turkish and Japanese. This demonstrates that open-source LLMs have rapidly closed much of the gap thanks to scale and improved training. However, there remains a gap in the truly low-resource domain —GPT-4o still had a clear edge there. Mistral Large 2 ( 128B) performed well but was slightly behind LLaMA-3.1 (70B), suggesting data and architecture differences. As of late 2025, closed models still lead particularly in nuanced cultural understanding.

- **Impact of language characteristics:** Turkish tested morphological generalization; smaller models often failed to produce correct case endings or plural suffixes. Even large models occasionally made minor errors, aligning with prior studies that LLMs are not fully systematic in morphology. Japanese tested politeness and context sensitivity —most models defaulted to polite forms, though only top models adjusted tone to context. Cantonese tested dialectal robustness and low-resource knowledge. GPT-4o, GPT-4, and Claude handled Cantonese reasonably well, but open models struggled, reflecting data scarcity. Training open models on Mandarin-dominant corpora limits their Cantonese ability. Future research could explore fine-tuning or prompt-based adaptation for dialects.

- **Cultural and contextual understanding:** Our culturally grounded dialogue evaluation suggests that state-of-the-art LLMs are beginning to grasp culture-specific context. GPT-4o showed superior adaptation to local norms (choice of words, humor). Nonetheless, even top models are imperfect and could benefit from more culturally diverse training or explicit alignment. Smaller models often defaulted to generic, culture-agnostic behavior, indicating that cultural sensitivity may emerge only beyond a certain model capacity and training diversity. This underscores the importance of including diverse cultural dialogues, idioms, and real-world conversational data in training. As noted by prior work [29, 38], the field lacks standardized evaluation for cultural intelligence; our benchmark is a step toward filling that gap.

- **Human vs. automated evaluation:** We observed cases where metrics and human judgments diverged —for example, ROUGE penalizing fluent paraphrases or BLEU underrating semantically correct translations. Human evaluation was crucial for true quality assessment, especially for dialogue where automatic metrics fail. LLM-as-a-judge approaches (e.g., GPT-4 scoring dialogues) show promise for scalable evaluation but risk bias if the judge shares the evaluated model's limitations. For low-resource languages, we recommend continuing to rely on human assessors, ideally native speakers, for trustworthy conclusions.

- **Benchmark construction reflections:** Building this benchmark revealed several practical challenges. For Cantonese, data had to be manually crafted or translated —a time-intensive process but essential for inclusivity. Ensuring parallel difficulty across languages is hard; thus, our analysis focuses on relative performance per language rather than absolute cross-language comparison. A possible future design is truly parallel multilingual datasets (same facts expressed in different languages), though that only suits factual QA tasks. In summarization and dialogue, we prioritized naturalness over strict parallelism, letting each language reflect its culture authentically.

All prompts and outputs will be released with this paper to foster transparency and further extension —adding more languages (e.g., African or South-Asian low-resource ones) or tasks (e.g., code-mixed dialogue, speech transcription). Expanding evaluations will further push LLM developers to cover all linguistic and cultural domains.

# 6 Conclusion

We presented a comprehensive cross-lingual benchmark evaluating modern large language models on **low-resource and morphologically rich languages**, focusing on Turkish, Japanese, and Cantonese. Through four diverse tasks—open-domain QA, summarization, English-to-X translation, and culturally grounded dialogue—we examined seven major LLMs including GPT-4o, GPT-4, Claude 3.5 Sonnet, LLaMA 3.1, Mistral Large 2, LLaMA-2 13B, and Mistral 7B.

Our results reveal that:

- GPT-4o currently achieves the strongest multilingual performance, particularly on Cantonese and culturally sensitive dialogue.
- GPT-4 and Claude 3.5 remain highly competitive, while LLaMA 3.1 (70B) and Mistral Large 2 approach parity on Turkish and Japanese.
- Smaller open models (13B / 7B) still trail significantly, underscoring the persistent gap in low-resource linguistic coverage.

We find that **morphological complexity and cultural nuance** remain difficult even for the largest models, and that human evaluation is indispensable for judging fluency and cultural appropriateness. By combining quantitative metrics with human judgments, this work offers one of the first holistic multilingual analyses of the latest generation of LLMs.

We release our benchmark data, prompts, and human evaluation guidelines to encourage further research on:

- Scaling LLM training to include more underrepresented languages and dialects.
- Designing culturally aware fine-tuning and evaluation strategies.
- Building adaptive tokenization and morphology-sensitive architectures.

Ultimately, advancing multilingual understanding will make language technology more equitable, ensuring that linguistic and cultural diversity are reflected in future generations of AI systems.

# Acknowledgments

We thank the native speakers and linguists who contributed to data annotation and evaluation. This work was supported in part by institutional research funding and computational resources provided by our affiliated organizations. All opinions expressed are those of the authors.